\documentclass[conference]{IEEEtran}
\IEEEoverridecommandlockouts
\usepackage{amsmath,amssymb,amsfonts}
\usepackage{algorithmic}
\usepackage{graphicx}
\usepackage{textcomp}
\usepackage{xcolor}
\usepackage{multirow}
\usepackage{floatrow}
\usepackage[style=ieee, maxnames=6, minnames=1, maxbibnames=6, backend=biber]{biblatex}
\begin{document}

\title{Prompt-and-Check: Using Large Language Models to Evaluate Communication Protocol Compliance in Simulation-Based Training\\
\thanks{Thanks to Singapore Maritime Institute (SMI)}
}

\author{\IEEEauthorblockN{Vishakha Lall}
\IEEEauthorblockA{\textit{Centre of Excellence in Maritime Safety} \\
\textit{Singapore Polytechnic}\\
Singapore \\
vishakha\_lall@sp.edu.sg}
\and
\IEEEauthorblockN{Yisi Liu}
\IEEEauthorblockA{\textit{Centre of Excellence in Maritime Safety} \\
\textit{Singapore Polytechnic}\\
Singapore \\
liu\_yisi@sp.edu.sg}
}

\maketitle

\begin{abstract}
Accurate evaluation of procedural communication compliance is essential in simulation-based training, particularly in safety-critical domains where adherence to compliance checklists reflects operational competence. This paper explores a lightweight, deployable approach using prompt-based inference with open-source large language models (LLMs) that can run efficiently on consumer-grade GPUs. We present Prompt-and-Check, a method that uses context-rich prompts to evaluate whether each checklist item in a protocol has been fulfilled, solely based on transcribed verbal exchanges. We perform a case study in the maritime domain with participants performing an identical simulation task, and experiment with models such as LLama 2 7B, LLaMA 3 8B and Mistral 7B, running locally on an RTX 4070 GPU. For each checklist item, a prompt incorporating relevant transcript excerpts is fed into the model, which outputs a compliance judgment. We assess model outputs against expert-annotated ground truth using classification accuracy and agreement scores. Our findings demonstrate that prompting enables effective context-aware reasoning without task-specific training. This study highlights the practical utility of LLMs in augmenting debriefing, performance feedback, and automated assessment in training environments.
\end{abstract}

\begin{IEEEkeywords}
Large Language Models, Prompt Engineering, Zero-shot Inference
\end{IEEEkeywords}

\section{Introduction}
Assessing adherence to communication protocols in high-stakes domains such as healthcare, aviation, industry, and maritime safety is challenging. While structured checklists guide actions in high-risk scenarios, verifying compliance from naturalistic communication or behaviour is labour-intensive, relying on expert review of transcripts, logs, or recordings. Automating this process could greatly improve post-incident reviews, training feedback, and safety audits.

LLMs have shown strong capabilities in a range of reasoning and classification tasks, especially when guided by natural language prompts. Prompt-based techniques have been widely explored for zero-shot and few-shot inference in tasks such as question answering, natural language inference \cite{brown2020languagemodelsfewshotlearners, sanh2022multitaskpromptedtrainingenables} and structured information extraction \cite{zhou2023leasttomostpromptingenablescomplex}. These approaches leverage instruction-tuned models to follow task-specific prompts without requiring task-specific fine-tuning. Recent work has also demonstrated the use of LLMs for procedural and checklist-based tasks in domains such as healthcare \cite{medicalllm} and aviation safety \cite{aviationllm}. However, these applications assume access to clean, well-structured input. The challenge of applying LLMs to noisy, multi-turn natural language transcripts for compliance classification remains underexplored.

This paper explores the use of prompt-based LLMs to infer communication checklist compliance from natural language communication transcripts. We present a generalisable methodology that combines temporal and semantic context selection with schema-constrained prompting, enabling LLMs to output structured compliance judgments and accompanying justifications. The approach is designed to be model-agnostic and compatible with local deployment on resource-constrained hardware. We evaluate three competitive, open-weight models, LLaMA 2 7B, LLaMA 3 7B, and Mistral 7B, on their ability to generate accurate, schema-compliant compliance decisions. 
This work demonstrates the feasibility of leveraging locally runnable LLMs for structured, interpretable assessment of human behaviour in protocol-governed tasks. As a representative use case, we apply our method to a structured maritime simulation dataset, where domain experts respond to critical scenarios. Each scenario has an expert-defined protocol checklist and corresponding communication transcripts. For each checklist item, we extract relevant transcript segments and prompt the LLM to assess compliance, producing both a classification label and a supporting justification.

\section{Dataset}
The dataset was collected at the Advanced Navigation Research Simulator (ANRS) at the Centre of Excellence in Maritime Safety (CEMS), Singapore. The controlled experiment evaluated how effectively experienced deck officers follow predefined safety procedures during emergency navigational events. Ten licensed deck officers, each with prior maritime navigation experience, completed two simulation trials, one in good visibility and one in poor visibility, yielding 20 sessions. Both trials were functionally identical in scenario content but differed in visual conditions, enabling analysis of protocol adherence consistency across environments. Each 45-minute trial required navigating through congested waters while encountering three pre-scripted high-risk events: potential vessel collision, engine failure, and severe storm. Event timing and nature were consistent across participants and conditions for comparability. For each event, maritime safety experts finalised an ordered protocol checklist specifying the expected verbal actions, decisions, and communications. These checklists, summarised in Table \ref{checklist}, served as the basis for assessing whether participants’ responses aligned with expected procedures. All verbal communications during the simulation were recorded and transcribed using a maritime-specific automated speech recognition (ASR) model \cite{lall2024contextualbiasingimprovedomainspecific}. Transcripts include time-ordered utterances from the participant and simulated entities (played by the instructor during the simulation), forming the core textual input for prompt-based evaluation. Each simulation thus provides: a transcript of participant communication ($T^{(i)}$), a scenario-specific checklist ($C^{(i)}$), and expert-annotated labels ($y^{(i)}$) indicating ground-truth compliance for each checklist item.

\begin{table}[]
\centering
\resizebox{0.8\textwidth}{!}{
\begin{tabular}{|llll|}
\hline
\begin{tabular}[c]{@{}l@{}}Injected\\ Scenario\end{tabular} & \begin{tabular}[c]{@{}l@{}}Visibility\\ Conditions\end{tabular} & Checklist Item & \begin{tabular}[c]{@{}l@{}}Ordered \\ priority\\ (1-lowest, \\ 4-highest)\end{tabular} \\ \hline
\multirow{3}{*}{\begin{tabular}[c]{@{}l@{}}Potential\\ collision \\ with a \\ nearby \\ vessel\end{tabular}} & \multirow{3}{*}{\begin{tabular}[c]{@{}l@{}}Daytime/\\ Nighttime\end{tabular}} & \begin{tabular}[c]{@{}l@{}}Report own vessel's current position \\ and heading to port control\end{tabular} & 4 \\ \cline{3-4} 
 &  & \begin{tabular}[c]{@{}l@{}}Verify and discuss unidentified \\ vessel with helmsman\end{tabular} & 3 \\ \cline{3-4} 
 &  & \begin{tabular}[c]{@{}l@{}}Check with port control regarding \\ the unidentified vessel\end{tabular} & 2 \\ \hline
\multirow{7}{*}{\begin{tabular}[c]{@{}l@{}}Main \\ Engine\\ Failure\end{tabular}} & \multirow{6}{*}{\begin{tabular}[c]{@{}l@{}}Daytime/\\ Nighttime\end{tabular}} & \begin{tabular}[c]{@{}l@{}}Contact engine room to know the \\ fault status and estimate time to \\ resolution\end{tabular} & 4 \\ \cline{3-4} 
 &  & Order anchoring stations on standby & 4 \\ \cline{3-4} 
 &  & Notify port control of engine failure & 4 \\ \cline{3-4} 
 &  & \begin{tabular}[c]{@{}l@{}}Inform port marine safety on engine \\ failure\end{tabular} & 3 \\ \cline{3-4} 
 &  & \begin{tabular}[c]{@{}l@{}}Broadcast engine failure and \\ reduced maneuverability to nearby \\ vessels\end{tabular} & 2 \\ \cline{3-4} 
 &  & Request tug assistance & 1 \\ \cline{2-4} 
 & Nightime & \begin{tabular}[c]{@{}l@{}}Issue command to display NUC \\ (Not Under Command) lights\end{tabular} & 2 \\ \hline
\multirow{5}{*}{\begin{tabular}[c]{@{}l@{}}Severe \\ Storm\end{tabular}} & \multirow{5}{*}{\begin{tabular}[c]{@{}l@{}}Daytime/\\ Nighttime\end{tabular}} & \begin{tabular}[c]{@{}l@{}}Contact bridge team to assign \\ lookouts\end{tabular} & 4 \\ \cline{3-4} 
 &  & Update engine room & 4 \\ \cline{3-4} 
 &  & \begin{tabular}[c]{@{}l@{}}Update port control on vessel \\ status and intention\end{tabular} & 4 \\ \cline{3-4} 
 &  & Update nearby vessels on position & 4 \\ \cline{3-4} 
 &  & Keep anchoring stations on standby & 3 \\ 
 \hline
\end{tabular}}
\caption{Safety protocol checklist for risky events with priority}
\label{checklist}
\end{table}

\section{Proposed Methodology}

\subsection{Task Framing} 
Let the simulation-based scenario be represented as a tuple,
\begin{equation}
    S=(T,C)
\end{equation}
where,
\begin{itemize}
    \item $T=\{t_1, t_2,..., t_n\}$ is the ordered set of all transcribed communication utterances during the scenario
    \item $C=\{c_1, c_2,..., c_m\}$ is the set of expected checklist items as part of the procedural protocol for the simulation scenario. Each checklist item $c_j \in C$ is a structured, task-relevant action or requirement.
\end{itemize} For each checklist item $c_j$, the goal is to determine its compliance status,
\begin{equation}
    y_j=f_\theta(T,c_j)\in\{True, False\}
\end{equation} where,
\begin{itemize}
    \item $y_j$ is the predicted compliance label
    \item $f_\theta$ is a language model-based function, parameterised by $\theta$ that uses prompt-based inference to make a decision using the input transcript $T$ and the checklist item $c_j$
\end{itemize} To formulate the function $f_\theta$ as a prompting operation,
\begin{equation}
    f_\theta(T,c_j) = LLM_\theta(Prompt(T,c_j))
\end{equation} where,
\begin{itemize}
    \item $LLM_\theta$ denotes the language model
    \item $Prompt(\cdot)$ is a function that generates a structured natural language prompt
\end{itemize} To ensure computational tractability and relevance, we define a context window $T_j \subseteq T$ for each $c_j$, which includes utterances semantically and temporally aligned with the checklist item,
\begin{equation}
    T_j = SelectContext(T, c_j)
\end{equation} Then,
\begin{equation}
    f_\theta(T,c_j) = LLM_\theta(Prompt(T_j,c_j))
\end{equation}

\subsection{Context Selection}
Accurate checklist compliance evaluation requires the language model to reason over only the transcript segments relevant to each checklist item. Supplying the full simulation transcript is computationally inefficient and may reduce accuracy due to irrelevant content. To address this, we use a two-stage context selection method that balances temporal precision with semantic relevance.
\subsubsection{Temporal Context Extraction}
Each simulation trial was instrumented with a predefined timeline of event injections, with associated start and end timestamps. These timestamps serve as coarse temporal anchors for isolating communication relevant to the scenario in which the checklist items are situated. For a checklist item $c_j$, associated with an injected event $e_k$, we define a primary context window,
\begin{equation}
    T_{e_k} = \{t_i \in T |t_{e_k}^{start} -\Delta_p <t_i<t_{e_k}^{end}+\Delta_f\}
\end{equation} where,
\begin{itemize}
    \item $t_{e_k}^{start}$ and $t_{e_k}^{end}$ are the start and end timestamps of injected event $e_k$
    \item $\Delta_p$ and $\Delta_f$ are pre- and post-buffers to capture surrounding context
    \item $ T_{e_k}$ is the temporally extracted canditate set of utterances
\end{itemize}
\subsubsection{Semantic Similarity-Based Refinement}
To further refine the rule-based context and focus on utterances most relevant to a specific checklist item $c_j$, we apply a semantic similarity filtering step. Each utterance $t_i \in T_{e_k}$ and the checklist item $c_j$ are embedded into a shared vector space using a sentence embedding model MiniLM \cite{wang2020minilmdeepselfattentiondistillation}. Let $\phi(t_i)$ and $\phi(c_j)$ be the embeddings of the utterance and checklist item, respectively. We compute the Cosine similarity as,
\begin{equation}
    sim(t_i, c_i) = \frac{\phi(t_i)\cdot\phi(c_i)}{||\phi(t_i)||\cdot||\phi(c_j)||}
\end{equation} We retain utterances $t_i$ where $sim(t_i, c_i) \textgreater \tau$, where $\tau = 0.7$ is an empirically derived threshold. 

The final context window $T_j$ is,
\begin{equation}
    T_j = \{t_i \in T_{e_k} | sim(t_i, c_j) > \tau \}
\end{equation} This dual filtering approach only processes context that is both temporally aligned and semantically relevant.

\subsection{Prompt Design}
Each prompt is constructed with three main components:
\begin{enumerate}
    \item Task Introduction: A clear and direct instruction that defines the goal
    \item Scenario Context: A filtered segment of the transcript, selected using the context selection methodology
    \item Target Checklist Item: A single checklist action under evaluation
\end{enumerate} An example of the prompt template is illustrated in Table \ref{prompt}.

\begin{table}[]
\resizebox{0.9\textwidth}{!}{
\begin{tabular}{|l|}
\hline
Task Introduction: \\
\begin{tabular}[c]{@{}l@{}}You are an assistant tasked with evaluating the maritime communication of a \\ participant attempting a simulated exercise. In this scenario, the participant is \\ being assessed on their ability to avoid potential collisions with nearby \\ vessels. You will be provided with the participant's transcript and the \\ checklist item. You are required to identify whether the checklist item was \\ explicitly addressed by the participant in the transcript. Return a JSON object \\ with the following keys:\\ is\_completed: True or False\\ index: If is\_completed is True, capture the timestamp of the transcript\\ utterance where the adherence was first found\\ evidence: A direct quote from the transcript as justification\end{tabular} \\ \hline
Scenario Context: \\
\begin{tabular}[c]{@{}l@{}}{[}\{index: 27, transcript: "Port Control, Port Control, this is Adventurer, we are \\ proceeding towards Eastern Boarding Ground Charlie and we have a vessel \\ crossing ahead of us. Can you give us the name of that vessel?"\}, \\ \{index: 32, transcript: "Challenger, Challenger, this is Adventurer, can you \\ please share your intention and heading?"\}{]}\end{tabular} \\ \hline
Target Checklist Item: \\
Report own vessel’s current position and heading to port control \\ \hline
\end{tabular}}
\caption{Sample Prompt}
\label{prompt}
\end{table}

\subsection{Schema Constrained Parsing and Validation}
To increase reliability and reduce hallucinated responses, we implement a $JSONSchemaParser$ that post-processes the raw output from the LLM. It performs structural validation to ensure JSON keys and values match the schema, type enforcement to flag invalid answer types, and fallback parsing to attempt auto-correction if the structure is violated.

\subsection{Models}
This study employs three state-of-the-art, open-weight LLMs, LLaMA 2 7B, LLaMA 3 8B, and Mistral 7B, that can run locally on a single NVIDIA RTX 4070 GPU. Selection criteria included strong reasoning performance, instruction-following capability, and schema-constrained output support.

The LLaMA 2 7B model \cite{touvron2023llama2openfoundation}, is a 32-layer, 7B-parameter decoder-only transformer trained on 2T+ tokens and fine-tuned for instruction following. It excels in factual recall and structured output for simpler tasks, serving as a solid baseline for prompt-based compliance detection.

LLaMA 3 8B \cite{grattafiori2024llama3herdmodels}, improves on its predecessor with a redesigned tokeniser, optimised attention, better training data, and an 8K-token context window—ideal for long multi-turn transcripts. Enhanced instruction tuning boosts reasoning and structured output for complex, temporally grounded scenarios.

The Mistral 7B model \cite{jiang2023mistral7b}, is a compact, high-efficiency transformer with grouped-query and sliding window attention for lower memory use and faster inference. Despite its size, it delivers strong structured generation performance, making it well-suited for real-time or resource-limited applications.

\subsection{Evaluation Metrics}
To evaluate prompt-based LLMs for safety checklist compliance, we use quantitative and qualitative metrics focused on two aspects: correctness in identifying adherence and the quality of justifications derived from natural language transcripts.

\subsubsection{Weighted Checklist Compliance Accuracy}
The proportion of checklist items for which the model's predicted compliance label (True, False) matches the ground truth annotation, weighted by their priority. 
\begin{equation}
    Accuracy = \frac{1}{N}\sum_{n=1}^Np_n(\hat{y_n} = y_n)
\end{equation} where,
\begin{itemize}
    \item $N$ is the total number of checklist item evaluations across all scenarios and participants
    \item $\hat{y_n}$ is the predicted label for the $n^{th}$ item
    \item ${y_n}$ is the annotated ground truth
    \item $p_n$ is the normalised priority of the checklist item, normalised by scenario
\end{itemize}

\subsubsection{Justification Alignment Score}
Each model output includes a natural language explanation citing transcript evidence. We assess justification quality using a manually rated Justification Alignment Score on a 3-point Likert scale: 2 (fully aligned, clearly references relevant transcript phrases), 1 (partially aligned, vague but generally consistent), and 0 (misaligned, irrelevant or contradictory). Average scores per model reflect their ability to produce meaningful, grounded rationales.

\section{Results}

Table \ref{results} summarises the comparative performance of LLaMA 2 7B, LLaMA 3 8B, and Mistral 7B across three representative simulation scenarios. We evaluate models using two primary metrics: Average Weighted Checklist Compliance Accuracy and Average Justification Alignment Score. LLaMA 3 8B outperforms the other models across all scenarios, achieving the highest overall compliance accuracy of $93.6\%$ and the highest overall justification alignment score of $1.8$. LLaMA 2 7B and Mistral 7B exhibit comparable performance. Across individual scenarios, all models perform best in the severe storm scenario, likely due to more explicit communication patterns. The potential collision scenario presents the most challenge, suggesting that nuanced situational cues like identification of an unidentified vessel are harder for models to capture without strong context filtering. Justification alignment scores closely follow accuracy trends, validating that correct predictions are generally supported by coherent rationale. 

Table \ref{ablation} presents an ablation study investigating how different context selection methods affect model performance. The baseline condition performs poorly, confirming that using the entire transcript without context refinement leads to overwhelming noise and misalignment. Temporal context extraction alone significantly improves performance. Semantic filtering in isolation offers moderate gains but is less effective than temporal filtering. The combined method, which first extracts temporal windows and then refines them using semantic similarity, achieves the best results across all models. Additionally, this method reduces the average number of input utterances to $6.1$, indicating a highly efficient and focused prompt structure.

\begin{table}[]
\centering
\resizebox{0.9\textwidth}{!}{
\begin{tabular}{|lllll|llll|}
\hline
\multirow{2}{*}{Model} & \multicolumn{4}{c|}{\begin{tabular}[c]{@{}c@{}}Average Weighted Checklist Compliance \\Accuracy\end{tabular}} & \multicolumn{4}{c|}{\begin{tabular}[c]{@{}c@{}}Average Justification Alignment Score \\ (0-misaligned, 2-fully aligned)\end{tabular}} \\ \cline{2-9} 
 & \begin{tabular}[c]{@{}l@{}}Potential \\ collision\\ with a \\ nearby\\ vessel\end{tabular} & \begin{tabular}[c]{@{}l@{}}Main\\ engine \\ failure\end{tabular} & \begin{tabular}[c]{@{}l@{}}Severe\\ storm\end{tabular} & Overall & \begin{tabular}[c]{@{}l@{}}Potential \\ collision\\ with a \\ nearby\\ vessel\end{tabular} & \begin{tabular}[c]{@{}l@{}}Main\\ engine \\ failure\end{tabular} & \begin{tabular}[c]{@{}l@{}}Severe\\ storm\end{tabular} & Overall \\ \hline
LLaMA 2 7B & 89.1 & 92.6 & 93.4 & 91.7 & 1.6 & 1.6 & 1.7 & 1.6 \\ \hline
LLaMA 3 8B & 91.7 & 94.3 & 94.8 & 93.6  & 1.7 & 1.8 & 1.8 & 1.8 \\ \hline
Mistral 7B & 88.8 & 92.2 & 92.7 & 91.2 & 1.5 & 1.6 & 1.6 & 1.6 \\
\hline
\end{tabular}}
\caption{Comparative metrics by scenario}
\label{results}
\end{table}

\begin{table}[]
\centering
\resizebox{0.9\textwidth}{!}{
\begin{tabular}{|l|l|ccc|ccc|}
\hline
\multirow{4}{*}{} & \multirow{4}{*}{\begin{tabular}[c]{@{}l@{}}Average number \\ of transcript\\ utterances in\\ scenario context\end{tabular}} & \multicolumn{3}{c|}{\begin{tabular}[c]{@{}c@{}}Average checklist compliance \\ accuracy\end{tabular}} & \multicolumn{3}{c|}{\begin{tabular}[c]{@{}c@{}}Average justification alignment\\ score\end{tabular}} \\ \cline{3-8} 
 &  & \multicolumn{1}{l}{\multirow{2}{*}{LLaMA 2 7B}} & \multicolumn{1}{l}{\multirow{2}{*}{LLaMA 3 8B}} & \multicolumn{1}{l|}{\multirow{2}{*}{Mistral 7B}} & \multicolumn{1}{l}{\multirow{2}{*}{LLaMA 2 7B}} & \multicolumn{1}{l}{\multirow{2}{*}{LLaMA 3 8B}} & \multicolumn{1}{l|}{\multirow{2}{*}{Mistral 7B}} \\
 &  & \multicolumn{1}{l}{} & \multicolumn{1}{l}{} & \multicolumn{1}{l|}{} & \multicolumn{1}{l}{} & \multicolumn{1}{l}{} & \multicolumn{1}{l|}{} \\
 &  & \multicolumn{1}{l}{} & \multicolumn{1}{l}{} & \multicolumn{1}{l|}{} & \multicolumn{1}{l}{} & \multicolumn{1}{l}{} & \multicolumn{1}{l|}{} \\ \hline
\begin{tabular}[c]{@{}l@{}}No context\\ selection\end{tabular} & \multicolumn{1}{c|}{37.8} & 24.3 & 26.9 & 23.6 & 0.7 & 0.8 & 0.7 \\ \hline
\begin{tabular}[c]{@{}l@{}}Temporal context \\ extraction only\end{tabular} & \multicolumn{1}{c|}{11.2} & 70.1 & 76.3 & 71.1 & 1.2 & 1.4 & 1.2 \\ \hline
\begin{tabular}[c]{@{}l@{}}Semantic context \\ extraction only\end{tabular} & \multicolumn{1}{c|}{19.5} & 58.3 & 58.7 & 57.9 & 1.1 & 1.3 & 1.1 \\ \hline
\begin{tabular}[c]{@{}l@{}}Temporal context \\ extraction followed\\ by semantic similarity\\  refinement\end{tabular} & \multicolumn{1}{c|}{6.1} & 90.7 & 93.6 & 91.2 & 1.6 & 1.8 & 1.6 \\
\hline
\end{tabular}}
\caption{Ablation over context selection methodologies}
\label{ablation}
\end{table}

\section{Conclusion}
This work presents a practical application of prompt-based large language models (LLMs) for structured protocol compliance assessment from naturalistic communication transcripts. By leveraging a combination of temporal and semantic context selection and structured prompting, we demonstrate that open-weight LLMs can reliably determine checklist adherence and provide aligned justifications.

Beyond the case study presented, this approach has broad potential in domains where verbal protocols or standard operating procedures are critical, such as aviation, emergency response, healthcare, and industrial safety. Future work may explore generalising across different types of protocols, incorporating multimodal inputs (e.g., video, sensor data), and extending this method for real-time decision support or debriefing tools.

\printbibliography
\end{document}